# Dr.Fill: Crosswords and an Implemented Solver for Singly Weighted CSPs

**Matthew L. Ginsberg**
*On Time Systems, Inc.*
*355 Goodpasture Island Road, Suite 200*
*Eugene, Oregon  97401*

## Abstract

We describe Dr.Fill, a program that solves American-style crossword puzzles. From a technical perspective, Dr.Fill works by converting crosswords to weighted csps, and then using a variety of novel techniques to find a solution. These techniques include generally applicable heuristics for variable and value selection, a variant of limited discrepancy search, and postprocessing and partitioning ideas. Branch and bound is not used, as it was incompatible with postprocessing and was determined experimentally to be of little practical value. Dr.Fill's performance on crosswords from the American Crossword Puzzle Tournament suggests that it ranks among the top fifty or so crossword solvers in the world.

## 1. Introduction

In recent years, there has been interest in solving constraint-satisfaction problems, or csps, where some of the constraints are "soft" in that while their satisfaction is desirable, it is not strictly required in a solution. As an example, if a construction problem is modeled as a csp, it may be possible to overutilize a particular labor resource by paying the associated workers overtime. While not the cheapest way to construct the artifact in question, the corresponding solution is certainly viable in practice.

Soft constraints can be modeled by assigning a cost to violating any such constraint, and then looking for that solution to the original csp for which the accumulated cost is minimized.

By and large, work on these systems has been primarily theoretical as various techniques for solving these "weighted" csps (wcsps) are considered and evaluated without the experimental support of an underlying implementation on a real-world problem. Theoretical complexity results have been obtained, and the general consensus appears to be that some sort of branch-and-bound method should be used by the solver, where the cost of one potential solution is used to bound and thereby restrict the subsequent search for possible improvements.

Our goal in this paper is to evaluate possible wcsp algorithms in a more "practical" setting, to wit, the development of a program (Dr.Fill) designed to solve American-style crossword puzzles. Based on the search engine underlying Dr.Fill, our basic conclusions are as follows:

1. We present specific variable- and value-selection heuristics that improve the effectiveness of the search enormously.

2. The most effective search technique appears to be a modification of limited discrepancy search (LDS) (Harvey & Ginsberg, 1995).

3. Branch-and-bound appears *not* to be a terribly effective solution technique for at least some problems of this sort.

4. Postprocessing complete candidate solutions improves the effectiveness of the search.

A more complete description of the crossword domain can be found in Section 2.2; example crosswords appear in Figures 1 and 2. The overall view we will take is that, given both a specific crossword clue $c$ and possible solution word or "fill" $f$, there is an associated score $p(f|c)$ that gives the probability that the fill is correct, given the clue. Assuming that these probabilities are independent for different clues, the probability that a collection of fills solves a puzzle correctly is then simply

$$\prod_i p(f_i|c_i) \tag{1}$$

where $f_i$ is the fill entered in response to clue $c_i$. DR.FILL's goal is to find a set of fills that is legal (in that intersecting words share a letter at the square of intersection) while maximizing (1).

For human solvers, $p(f|c)$ will in general be zero except for a handful of candidate fills that conform to full domain knowledge. Thus a "1973 nonfiction best seller about a woman with multiple personalities" must be "Sybil"; a 3-letter "Black Halloween animal" might be "bat" or "cat", and so on. For DR.FILL, complete domain knowledge is impractical and much greater use is made of the crossing words, as the CSP solver exploits the hard constraints in the problem to restrict the set of candidate solutions.

DR.FILL's performance as a solver is comparable to (but significantly faster than) all but the very best human solvers. In solving New York Times crosswords (which increase in difficulty from Monday to Saturday, with the large Sunday puzzles comparable to Thursdays in difficulty), DR.FILL generally solves Monday to Wednesday puzzles fairly easily, does well on Friday and Saturday puzzles, but often struggles with Thursday and Sunday puzzles. These puzzles frequently involve some sort of a "gimmick" where the clues or fill have been modified in some nonstandard way in order to make the puzzle more challenging. When run on puzzles from the American Crossword Puzzle Tournament, the annual national gathering of top solvers in the New York City area, DR.FILL's performance puts it in the top fifty or so of the approximately six hundred solvers who typically attend.

The outline of this paper is as follows. Preliminaries are contained in the next section, both formal preliminaries regarding CSPs in Section 2.1 and a discussion of crosswords in Section 2.2. Section 2.3 discusses crosswords as CSPs specifically, including a description of the variety of ways in which crosswords differ from the problems typically considered by the constraint satisfaction community.

The heuristics used by DR.FILL are described in Section 3, with value-selection heuristics the topic of Section 3.1 and variable-selection heuristics the topic of Section 3.2. The techniques for both value and variable selection can be applied to WCSPs generally, although it is not clear how dependent their usefulness is on the crossword-specific features described in Section 2.3.

Our modification of LDS is described in Section 4, and is followed in Section 5 with our first discussion of the experimental performance of our methods. Algorithmic extensions involving postprocessing are discussed in Section 6, which also discusses the reasons that branch-and-bound techniques are not likely to work well in this domain. Branch-and-bound and postprocessing are not compatible but the arguments against branch-and-bound are deeper than that. Section 7 describes the utility of splitting a crossword into smaller problems when the associated constraint graph disconnects, an idea dating back to work of Freuder and Quinn (1985) but somewhat different in the setting provided by LDS.

Section 8 concludes by describing related and future work, including the earlier crossword solvers PROVERB (Littman, Keim, & Shzaeer, 2002) and WEBCROW (Ernandes, Angelini, & Gori, 2005), and the Jeopardy-playing program WATSON (Ferrucci, Brown, Chu-Carroll, Fan, Gondek, Kalyanpur, Lally, Murdock, Nyberg, Prager, Schlaefer, & Welty, 2010).

## 2. Preliminaries

In this section, we give a brief overview of constraint satisfaction, crossword puzzles, and the relationship between the two.

### 2.1 Constraint Satisfaction

In a conventional constraint-satisfaction problem, or CSP, the goal is to assign values to variables while satisfying a set of constraints. The constraints indicate that certain values for one variable, say $v_1$, are inconsistent with other specific values for a different variable $v_2$.

Map coloring is a typical example. If a particular country is colored red, then neighboring countries are not permitted to be the same color.

We will formulate crossword solving by associating a variable to each word in the crossword, with the value of the variable being the associated fill. The fact that the first letter of the word at 1-Across has to match the first letter of the word at 1-Down corresponds to a constraint between the two variables in question.

A basic CSP, then, consists of a set $V$ of variables, a set $D$ of domains, one for each variable, from which the variables' values are to be taken, and a set of constraints.

**Definition 2.1** *Given a set of domains $D$ and set $V$ of variables, an n-ary constraint $\lambda$ is a pair $(T, U)$ where $T \subseteq V$ is of size n and $U$ is a subset of the allowed sets of values for the variables in $V$. An assignment $S$ is a mapping from each variable $v \in V$ to an element of $v$'s domain. For $T \subseteq V$, the restriction of $S$ to $T$, to be denoted $S|_T$, is the restriction of the mapping $S$ to the set $T$. We will say that $S$ satisfies the constraint $(T, U)$ if $S|_T \in U$.*

The constraint simply specifies the sets of values that are allowed for the various variables involved.

As an example, imagine coloring a map of Europe using the four colors red, green, blue and yellow. Now $T$ might be the set {France, Spain} (which share a border), $D$ would assign the domain {red, green, blue, yellow} to each variable, and $U$ would be the twelve ordered pairs of distinct colors from the four colors available. The associated constraint indicates that France and Spain cannot be colored the same color.

As we have remarked, we will take the view that the variables in a crossword correspond to the various slots into which words must be entered, and the values to be all of the words in some putative dictionary from which the fills are taken (but see the comments in Section 2.3). If two words intersect (e.g., 1-Across and 1-Down, typically), there is a binary constraint excluding all pairs of words for which shared letters differ.

**Definition 2.2** *A* constraint-satisfaction problem, *or* CSP, *is a triple* $(V, D, \Lambda)$ *where* $V$ *is a set of variables,* $D$ *gives a domain for each variable in* $V$, *and* $\Lambda$ *is a set of constraints.* $|V|$ *will be called the* size *of the* CSP. *If every constraint in* $\Lambda$ *is either unary or binary, the* CSP *is called a* binary CSP.

*For a* CSP $C$, *we will denote the set of variables in* $C$ *by* $V_C$, *the domains by* $D_C$, *and the constraints by* $\Lambda_C$.

*A* solution *to a* CSP *is an assignment that satisfies every constraint in* $\Lambda$.

Both map coloring and crossword solving as described above are binary CSPs.

There is an extensive literature on CSPs, describing both their applicability to a wide range of problems and various techniques that are effective in solving them. It is not practical for me to repeat that literature here, but there are two points that are particularly salient.

First, CSPs are generally solved using some sort of backtracking technique. Values are assigned to variables; when a conflict is discovered, a backtrack occurs that is designed to correct the source of the problem and allow the search to proceed. As with other chronologically based backtracking schemes, there are well-known heuristics for selecting the variables to be valued and the values to be used, and the most effective backtracking techniques use some kind of nogood reasoning (Doyle, 1979; Ginsberg, Frank, Halpin, & Torrance, 1990, and many others) to ensure that the backtrack will be able to make progress.

We will need to formalize this slightly.

**Definition 2.3** *Let* $C$ *be a* CSP, *and suppose that* $v$ *is a variable in* $V_C$ *and* $x$ *a value in the associated domain in* $D_C$. *By* $C|_{v=x}$ *we will denote the* CSP *obtained by setting* $v$ *to* $x$. *In other words,* $C|_{v=x} = (V_C - v, D_C, \Lambda)$ *where* $\Lambda$ *consists of all constraints* $\lambda$ *such that for some constraint* $(T, U) \in \Lambda_C$

$$\lambda = \begin{cases} (T, U), & \textit{if } v \notin T; \\ (T - v, \{u \in U | u(v) = x\}|_{T-v}), & \textit{if } v \in T. \end{cases} \quad (2)$$

$C|_{v=x}$ *will be called the* restriction *of* $C$ *to* $v = x$.

The notation may be intimidating but the idea is simple: Values permitted by constraints in the new problem are just those permitted in the old problem, given that we have decided to set $v$ to $x$. So if an original constraint doesn't mention $v$ (the top line in (2)), the new constraint is unchanged. If $v$ is mentioned, we see which values for the other variables are allowed, given that $v$ itself is known to take the value $x$.

**Definition 2.4** *Let $S$ be a partial solution to a* CSP $C$*, in that $S$ maps some of the variables in $V_C$ to elements of $D_C$. The* restriction *of $C$ to $S$, to be denoted $C|_S$, is the* CSP *obtained by successively restricting $C$ to each of the assignments in $S$ as in Definition 2.3.*

This definition is well-defined because we obviously have:

**Lemma 2.5** *The restriction defined in Definition 2.4 is independent of the order in which the individual restrictions are taken.* □

Backtracking CSP solvers work by selecting variables, trying various values for the variables so selected, and then recursively solving the restricted problems generated by setting the variable in question to the value chosen.

The second general point we would like to make regarding CSP solvers is that most implementations use some kind of forward checking to help maintain consistency as the search proceeds. As an example, suppose that we are about to assign a value $x$ to a variable $v$, but that if we do this, then every possible value for some other variable $v'$ will be eliminated by the constraints. In this case, we can obviously eliminate $x$ as a value for $v$.

It is worth formalizing this a bit, although we do so only in the most general of terms.

**Definition 2.6** *A* propagation mechanism $\pi$ *is a mapping from* CSP*s to* CSP*s that does not change the variables, so that $\pi(V, D, \Lambda) = (V, D', \Lambda')$ for any $(V, D, \Lambda)$. We also require that for any variable in $V$, the associated domain in $D'$ is a subset of the associated domain in $D$, and if $\lambda = (T, U) \in \Lambda$, there must be a $\lambda' = (T', U') \in \Lambda'$ with $U' \subseteq U$. We will say that $\pi$ is* sound *if, for any* CSP $C$ *and solution $S$ to $C$, $S$ is also a solution to $\pi(C)$.*

The propagation mechanism strengthens the constraints in the problem, and may reduce some of the variable domains as well. It is sound if it never discards a solution to the original CSP.

A wide range of propagation mechanisms has been discussed in the literature. Simplest, of course, is to simply eliminate variable values that can be shown to violate one of the constraints in the problem. Iterating this idea recursively until quiescence (Mackworth, 1977) leads to the well known AC-3 algorithm, which preserves *arc consistency* as the CSP is solved.

Returning to the map of Europe, Germany borders France, Holland, Poland, and Austria (among other countries). If Holland is red, Poland is blue, and Austria is yellow, this is sufficient to cause Germany's live set to be just green (assuming that these are the four colors available), which will in turn cause France's live set to exclude green, even though France does not share a direct constraint with Holland, Poland or Austria.

Alternatively, consider the crossword showed in Figure 1; this is the *New York Times* crossword (with solution) from March 10, 2011. Once we decide to put READING at 1-Across [Poet's performance], the live set for words at 1-Down consists only of six-letter words beginning with R. If we had also entered ASAMI at 21-Across ["Me, too"] and REDONDO at 27-Across, then the live set for 1-Down would be words of the form R...AR.

**Weighted CSPs** Sometimes it is desirable for a CSP to include "soft" constraints. The notion here is that while a soft constraint indicates a set of variable values that are preferred in a solution, the requirement is like the pirate's code, "more what you'd call 'guidelines'

**NY Times, Thu, Mar 10, 2011** Matt Ginsberg / Will Shortz

The completed crossword grid reads:

| R | E | A | D | I | N | G | ■ | C | E | L | L | A | R |
| A | L | B | A | N | I | A | ■ | S | O | M | E | O | N | E |
| W | O | R | D | S | P | R | O | N | O | U | N | C | E | D |
| B | I | O | ■ | S | Y | B | I | L | ■ | D | A | M | E |
| A | S | A | M | I | ■ | I | T | E | M | ■ | L | O | Y |
| R | E | D | O | N | D | O | ■ | R | A | P | I | N | E |
| ■ | B | A | I | L | E | Y | ■ | N | O | T | E | S |
| ■ | D | I | F | F | E | R | E | N | T | L | Y | ■ |
| S | P | E | L | L | ■ | A | N | G | E | L | I |
| C | A | M | E | A | S | ■ | G | O | E | S | B | A | D |
| R | D | A | ■ | T | O | R | I | ■ | S | H | A | R | E |
| A | T | R | A | ■ | M | I | S | S | M | ■ | N | I | A |
| W | H | E | N | C | A | P | I | T | A | L | I | Z | E | D |
| L | A | S | T | O | N | E | ■ | I | M | A | N | A | G | E |
| S | I | T | S | B | Y | ■ | R | A | I | N | I | E | R |

© 2011, The New York Times

## ACROSS

1. *Poet's performance
8. Frequent flooding site
14. Country with which the U.S. goes to war in "Wag the Dog"
15. Who "saved my life tonight" in a 1975 Elton John hit
16. With 36- and 58-Across, what the answers to the starred clues are
18. Jacket material, for short?
19. 1973 nonfiction best seller about a woman with multiple personalities
20. Lady of the knight?
21. "Me, too"
24. Line ___
26. "The Thin Man" actress
27. ___ Beach, Calif.
30. Plunder
32. Big name in circuses
35. B, A, D, G and E, e.g.
36. See 16-Across
38. Say "B-A-D-G-E," e.g.
40. Figures on the ceiling of la Cappella Sistina
41. Impersonated at a costume party
43. Spoils
47. Nutritional amt.
48. Doughnuts, but not danishes
51. Piece of the action
52. Gillette offering
54. Bette's "Divine" stage persona
57. Actress Vardalos
58. See 16-Across
62. "I'm done after this"
63. "Somehow everything gets done"
64. Does nothing
65. *Like Seattle vis-à-vis Phoenix

## DOWN

1. Seafood lover's hangout
2. Nancy Drew's aunt
3. One way to travel or study
4. Pop
5. Connections
6. Cheese ___
7. Player of golf
8. Clink
9. Prey of wild dogs and crocodiles
10. Furnish
11. Neighborhood
12. Flower that shares its name with a tentacled sea creature
13. They might depart at midnight
15. Huff
17. Japanese band
22. *Not fixed
23. Like Elgar's Symphony No. 1
25. Cloaks
28. "What's the ___?"
29. Pharmaceutical oils
31. *Shine
33. Old World eagle
34. Burglar in detective stories
35. William who played Uncle Charley on "My Three Sons"
37. Prefix with paganism
38. Many signatures
39. Noodle dish
42. Lots and lots of
44. Battle cry
45. French department in the Pyrenees
46. Less lively
49. Opportune
50. "Whatever it ___ don't care!"
53. Drones, maybe
55. Excitement
56. ___ Bear
59. Inner ear?
60. Medieval French love poem
61. What a keeper may keep

Figure 1: A Thursday *New York Times* crossword. ©2011 *The New York Times*. Reprinted with permission.

than actual rules." If no solution can be found without violating one of the soft constraints, it is acceptable to return a solution that does violate a soft constraint. The "hard" constraints are required to be satisfied in any case.

There are a variety of ways to formalize this. One of the simplest is to simply associate a cost with each soft constraint and to then search for the overall assignment of values to variables for which the total cost is minimized. We will take the view that the total cost of an assignment is the sum of the costs of the soft constraints that have been violated, although other accumulation functions (e.g., maximum) are certainly possible (Bistarelli, Montanari, Rossi, Schiex, Verfaillie, & Fargier, 1999).

A soft $k$-ary constraint thus consists of a mapping $c : D^k \to \mathbb{R}$ giving the cost associated with various selections for the variables being valued. The cost of a complete assignment of values to variables is the sum of the costs incurred for each soft constraint. A CSP including costs of this form is called a *weighted* constraint satisfaction problem, or WCSP (Larrosa & Schiex, 2004, and many others).

**Definition 2.7** *A weighted* CSP, *or* WCSP, *is a quadruple* $C = (V, D, \Lambda, W)$ *where* $(V, D, \Lambda)$ *is a* CSP *and* $W$ *is a set of pairs* $(U, c)$ *where* $U \subseteq V$ *is a set of variables and* $c$ *is a cost function assigning a cost to each assignment of the variables in* $U$. *Each element* $w \in W$ *will be called a* weighted constraint. *Where no ambiguity can arise, we will abuse notation and also denote by* $w$ *the associated cost function* $c$.

*Given a partial solution* $S$, *the associated* cost *of the weighted constraint* $w = (U, c)$, *to be denoted by* $c(S, w)$, *is the minimum cost associated by* $c$ *to any valuation for the variables in* $U$ *that extends the partial solution* $S$. *The* cost *of the partial solution is defined to be*

$$c(S) = \sum_{w \in W} c(S, w) \tag{3}$$

Informally, $c(S, w)$ is the minimum cost that will be charged by $w$ to any solution to $C$ that is an extension of $S$. We therefore have:

**Lemma 2.8** *Given a* WCSP $C$ *and partial solution* $S$, *every solution to* $C$ *that extends* $S$ *has cost at least* $c(S)$.  □

Note that our Definition 2.7 is slightly nonstandard in that we explicitly split the hard constraints in $\Lambda$ from the soft constraints in $W$. We do this because in the crossword domain, there is a further condition that is met: The soft constraints are always unary (although the hard constraints are not). There is simply a cost associated with setting the variable $v$ to some specific value $x$. We will refer to such problems as *singly* weighted CSPs, or SWCSPs. While the algorithmic ideas that we will present in this paper can be applied reasonably easily to WCSPs that are not in fact SWCSPs, the experimental work underlying Dr.Fill clearly reflects performance on SWCSPs specifically.[1]

As it is possible to use propagation to reduce the sizes of the domains in any particular CSP, it is also possible to use a variety of polynomial time algorithms to compute lower bounds on $c(S)$ for any partial solution $S$. The techniques used here have become increasingly sophisticated in recent years, ranging from algorithms that "move" costs around the constraint graph to better compute the minimum (Givry & Zytnicki, 2005; Zytnicki, Gaspin, de Givry, & Schiex, 2009) to more sophisticated approaches that solve linear programming problems to compute more accurate bounds (Cooper, de Givry, Sanchez, Schiex, Zytnicki, & Werner, 2010).

---

1. And in fact, Larrosa and Dechter (2000) have shown that all weighted CSPs can be recast similarly, into a form with only hard binary constraints and soft unary constraints.

Finally, we note in passing the every CSP has a "dual" version where the roles of the variables and constraints are exchanged. We view crosswords as CSPs where the variables are word slots and the values are the words that fill them, with constraints that require that the letters match where the words cross. But we could also view crosswords as CSPs where the variables are individual letters, the values are the usual A through Z, and the constraints indicate that every collection of letters needs to make up a legal word. We will discuss the likely relative merits of these two approaches in Section 2.3, after we have described crosswords themselves.

## 2.2 Crosswords

Since the introduction of the first "word cross" in the Sunday *New York World* almost a century ago (December 21, 1913), crosswords have become one of the world's most popular mental pastimes. Will Shortz, editor of the crossword for the *New York Times*, estimates that some five million people solve the puzzle each day, including syndication.[2]

### 2.2.1 Features of Crosswords

A typical *New York Times* crossword appears in Figure 1. We assume that the reader is familiar with the basic format, but there are many specific features that are worth mentioning.

**Crosswords are symmetric.** The black squares, or blocks, are preserved under a 180° rotation.[3] In addition, crosswords are almost always square in shape, with the *Times* daily puzzles being of size $15 \times 15$ and the Sundays $21 \times 21$.[4]

**Multiple words are permitted as fill.** In the puzzle of Figure 1, we have [Seafood lover's hangout] cluing RAW BAR at 1-Down and ["Somehow everything gets done"] cluing I MANAGE at 63-Across. There is no indication in the clue that a multiword answer is expected.

**Without the clues, crossword solutions are not unique.** There are many ways to fit words into any particular crossword grid; it is the clues that determine which legal fill is the puzzle's solution. This is what makes solving so challenging from a computational perspective: Failing to understand the clues (at least at some level) leaves the problem underconstrained.

---

2. Personal communication.

3. In rare cases, horizontal symmetry is present instead of rotational symmetry. In rarer cases still, the symmetry requirement is not honored.

4. The Sunday puzzles used to be $23 \times 23$ on occasion, but a reduction in the size of the *Times*' printed magazine section made these larger puzzles impractical.

**Puzzles can be themed or themeless.** A themeless puzzle contains a collection of generally unrelated words and clues. A themed puzzle has some shared element that connects many of the answers; when this happens, the shared answers are generally located symmetrically in the grid.

The puzzle in Figure 1 is themed. The (symmetric) entries at 1-Across, 65-Across, 22-Down and 31-Down are marked with asterisks and are all words that are pronounced differently when capitalized. This description also appears in the puzzle itself using the (also symmetric) entries at 16-Across, 36-Across and 58-Across.[5]

The presence of a theme has a profound impact on the solving experience. In this particular example (which is relatively straightforward), there are two entries (WORDSPRO-NOUNCED and WHENCAPITALIZED) that surely do not appear in any "dictionary" that the solver is using to fill the grid. There are also complex relationships among many of the entries – the phrase in 16-Across, 36-Across and 58-Across, but also the relationship of that phrase to the entries marked with asterisks.

Other themes present other challenges. Some of the more popular themes are *quip* puzzles, where a famous saying is split into symmetric pieces and inserted into the grid, and *rebus* puzzles, where more than one letter must be put in a single square. Arguably the most famous *Times* puzzle appeared on election day in 1996. The clue for 39-Across was [Lead story in tomorrow's newspaper (!), with 43-Across]. 43-Across was ELECTED and, depending on the choices for the down words, 39-Across could be either CLINTON or BOBDOLE. For example, 39-Down, a three-letter [Black Halloween animal] could be CAT (with the C in CLINTON) or BAT (with the B in BOBDOLE). So here, not only are there multiple ways to insert words legally into the grid, there are multiple ways for those words to match the clues provided. (Until the winner of the election was decided, of course.) The two legal solutions were identical except for the CLINTON/BOBDOLE ambiguity and the associated crossing words; it is not known whether there is a puzzle that admits multiple solutions without shared words, while conforming to the usual restrictions on symmetry, number of black squares, and so on.

A more extreme example of a themed crossword appears in Figure 2. Each clue has a particular letter replaced with asterisks (so in 1-A, for example, the e in [Twinkle] has been replaced). The letter that has been replaced is dropped in the fill, but the result is still a word. So 1-A, which would normally be GLEAM, becomes the David Bowie rock genre GLAM.

Every word in this puzzle is missing letters in this fashion. A computer (or human) solver will be unable to get a foothold of any kind if it fails to understand the "gimmick", and Dr. Fill fails spectacularly on this puzzle.

**Puzzles have structural restrictions.** For the *Times*, a daily unthemed puzzle will have at most 72 words; a themed puzzle will generally have at most 78. For a Sunday, 140 words is the limit. At most $\frac{1}{6}$ of the squares in the grid can be black squares.[6] This information is potentially of value to a solver because the word count can often be used to

---

**NY Times, Sun, May 17, 2009 TAKEAWAY CROSSWORD (see Notepad)** Matt Ginsberg / Will Shortz

© 2009, The New York Times

**ACROSS**

1. Twinkl*
5. Ou*look
9. "Dani*l Boon*" actor
13. Gung-*o [Lat.]
14. Sp*tted cats
15. Male chauvini*t, *ay
16. Playin* the slots, e.*.
17. Minia*ure desser*s
18. Admonis* [suffix]
19. *iding, as a sword
21. Neither young nor ol*
23. Diat*ibe delive*e*s
25. *hief
26. Sergea*t of old TV
29. Denta* devices
31. Still feelin* sleepy
32. *over up
34. Mentalist inspired by "Mandra*e the Magician"
38. Struc* with the foot
40. Mail origi*ator
42. Absolutely ama*e
43. *emoves pencil ma*ks
45. Big pi*kles?
47. Cat*' warning*
48. Apo*tle known a* "the Zealot"
50. Dise*se tr*nsmitted by cont*min*ted w*ter
52. Italian po*t?
55. Cerea* protein [Ger.]
57. P*blic sales
59. Im*roved one's lot [hyph.]
63. Can*ne restra*nt
64. Sou*hwes* German ci*y
66. Rea*y to be ri**en
67. Roun*e* up
68. Lessons fro* fables
69. 1960* prote*t [2 wds.]
70. S*ut up
71. Places *o pick up chicks?
72. Likel* to rise?

**DOWN**

1. Brib*, informally [Fr.]
2. Bit of *ire
3. Asla*'s realm
4. Lite*ally, "sac*ed utte*ances"
5. Nothing speci*l
6. Opposite of *on't
7. M*nhole em*n*tion
8. The *eatles' Paul McCartney, e.g.
9. Made smalle*
10. *ost co*prehensive
11. *mall amount, briefly
12. Device that re*oves stalks fro* fruit
14. Chin*s* cuisin*
20. *atchy *ony
22. *oting booth feature
24. Big name in ba**s
26. Pulit*er, e.g. [Fr.]
27. *egis*ative routine [Sp.]
28. B*ontë family
30. Speaks gi**erish
33. First mo*th, to Jua*
35. Lesley of "60 Minu*es"
36. Waiti*g o*e's tur*
37. Di*dainful expre**ion
39. Left the Met*oline*
41. Core cont*iners
44. National park in sout*west Tennessee
46. Turt*es and bu**ets have them
49. *mpos*ng house
51. Biase*
52. Qi*g Dy*asty people
53. A**ow shoote*s
54. O*erdo the diet
56. Street art, *aybe
58. Stron* *rowth
60. Not homo*eneous [Lat.]
61. "Lo*e Me Tender" star
62. B*rth cert., for one
65. Rank*es

Figure 2: A difficult themed crossword. ©2009 *The New York Times.* Reprinted with permission.

determine whether or not a puzzle has a theme. (The themed puzzle of Figure 1 has only 70 words, however, so all that can really be concluded is that a $15 \times 15$ puzzle with more than 72 words is likely themed in some way.)

There are other restrictions as well. Two-letter words are not permitted, and every square must have two words passing through it (alternatively, one-letter words are not permitted, either). The puzzle must be connected (in that the underlying CSP graph is as well). For a puzzle to be singly connected, in that converting a single empty square to a block disconnects it, is viewed as a flaw but an acceptable one.

**Fill words may not be repeated elsewhere in the puzzle.** If the fill BAT appears in one location in the puzzle, then it cannot be used elsewhere (including in multiword fill). In addition, BAT cannot be used in a clue. This means that words appearing in clues do not (or at least, should not) appear in the fill as well.

**Crossword clues must pass the "substitution test".** This is arguably the most important requirement from the solver's perspective. It must be possible to construct a sentence in which the clue appears, and so that the meaning of the sentence is essentially unchanged if the clue is replaced with the fill. So for the puzzle in Figure 1, one might say, "I've seen a video of e.e. cummings giving a READING," or the equivalent (although stilted) "I've seen a video of e.e. cummings giving a [poet's performance]." One might say, "We don't keep food in our house's CELLAR because we don't want it to get wet," or the equivalent "We don't keep food in our house's [frequent flooding site] because we don't want it to get wet."

The fact that the clues and associated fill must pass the substitution test means that it is generally possible to determine the part of speech, number (singular vs. plural) and tense (present, past, future, etc.) of the fill from the clue. So in Figure 1, the fill for 1-A is a singular noun, and so on. This restricts the number of possible values for each word considerably.

**There are other conventions regarding cluing.** If a clue contains an abbreviation, then the answer is an abbreviation as well. The puzzle in Figure 1 has no abbreviations clued in this way (a rarity), but 62-D in Figure 2 is [B*rth cert., for one]. The solution is IDENT, which is an abbreviation for "identification." Of course, the I gets dropped, so it is DENT that is entered into the grid. Abbreviations can also be indicated by a phrase like "for short" in the clue.

Clues ending in a ? generally indicate that some sort of wordplay is involved. In Figure 1, for example, we have 18-A: [Jacket material, for short?] The solution is BIO because "jacket" in the clue refers to a book's jacket, not clothing.

The wordplay often exploits the fact that the first letter of the clue is capitalized. So in Figure 1, 7-D is [Player of golf], referring to GARY Player. If the clue had been [Golf's Player], the capitalization (not to mention the phrasing) would have made it obvious that a proper name was involved. As the clue was written, the solver might easily be misled.

**Crossword features and Dr.Fill** Many of the features that we have described (e.g., symmetry) do not bear directly on the solving experience, and Dr.Fill is therefore unaware

---

repeated on August 7, 2010 by Joe Krozel. It is not known whether a 17-block puzzle of reasonable quality exists.

of them. The program does look for multiple-word fill and has a module that is designed to identify rebus puzzles. It does not check to see if fill words are repeated elsewhere, since this is so rare as to offer little value in the search. It uses fairly straightforward part-of-speech analysis to help with the substitution test, and checks clues for abbreviations. DR.FILL has no knowledge of puns.

## 2.3 Crossword Puzzles as SWCSPs

Given all of this, how are we to cast crossword solving as a CSP?

The view we will take, roughly speaking, is that we start with some large dictionary of possible fills, and the goal is to enter words into the grid that cross consistently and so that each word entered is a match for the associated clue. If $D$ is our dictionary, we will define $D_n$ to be the subset of $D$ consisting of words of exactly $n$ letters, so that BAT is in $D_3$, and so on. We also assume that we have a scoring function $p$ that scores a particular word relative to a given clue, which we will interpret probabilistically. Given a clue like [Black Halloween animal]$_3$ for a 3-letter word and potential fill BAT, $p(\text{BAT}|[\text{Black Halloween animal}]_3)$ is the probability that BAT is the correct answer for the word in question. Our goal is to now find the overall fill with maximum probability of being correct.

In other words, if $c_i$ is the $i$th clue and $f_i$ is the value entered into the grid, we want to find $f_i$ that satisfy the constraints of the problem (crossing words must agree on the letter filled in the crossing square) and for which

$$\prod_i p(f_i|c_i) \tag{4}$$

is maximized. As mentioned in the introduction, this assumes that the probabilities of various words being correct are uncorrelated, which is probably reasonably accurate but not completely correct in a themed puzzle.

If we define $\rho(f_i, c_i) = -\log p(f_i|c_i)$, maximizing (4) is equivalent to minimizing

$$\sum_i \rho(f_i, c_i) \tag{5}$$

This is exactly the SWCSP framework that we have described.

The dictionary $D_i$ must include not just words of length $i$, but also all sequences of words that are collectively of length $i$. (In other words, $D_{15}$ needs to include WHENCAPITALIZED.) In actuality, however, even this is not enough. There are many instances of crossword fill that are not even word sequences.

This may be because a "word" does not appear in any particular dictionary. A puzzle in the 2010 American Crossword Puzzle Tournament (ACPT) clued MMYY as [Credit card exp. date format] although MMYY itself is not a "word" in any normal sense. A further example is the appearance of SNOISSIWNOOW in the *Times* puzzle of 11/11/10, clued as "Apollo 11 and 12 [180 degrees]". Rotating the entire puzzle by 180 degrees and reading SNOISSIWNOOW upside down produces MOONMISSIONS.

Given these examples and similar ones, virtually *any* letter sequence can, in fact, appear in any particular puzzle. So the domain $D$ should in fact consist of all strings of the appropriate length, with the cost function used to encourage use of letter strings that are dictionary words (or sequences of words) where possible. This means that the variable domains are so large that both they and the associated $\rho$ functions must be represented functionally; computing either the dictionaries or the scores in their entirety is simply impractical.

This is the fundamental difference between the problem being solved by Dr.Fill and the problems generally considered in the AI literature. The number of variables is modest, on the order of 100, but the domain size for each variable is immense, $26^5$ or approximately 12 million for a word of length five, and some $1.7 \times 10^{18}$ for a word of length fifteen.

One immediate consequence of this is that Dr.Fill can do only a limited amount of forward propagation when it solves a particular puzzle. When a letter is entered into a particular square of the puzzle, it is effective to see the way in which that letter constrains the choices for the crossing words. But it appears *not* to be effective to propagate the restriction further. So if, in the puzzle in Figure 1, we restrict 1-Down to be a word of the form R...AR and the only word of that form in the dictionary is RAW BAR, we could conceivably then propagate forward from the B in BAR to see the impact on 18-Across. In actuality, however, the possibility that 1-Down be non-dictionary fill causes propagation beyond a simple one-level lookahead to be of negative practical value. The sophisticated propagation techniques mentioned in Section 2.1 appear not to be suitable in this domain.

A second consequence of the unrestricted domain sizes is that it is always possible to extend a partial solution in a way that honors the hard constraints in the problem. We can do this by simply entering random letters into each square of the puzzle (but only one letter per square, so that the horizontal and vertical choices agree). Each such random string is legal, and may even be correct. The reason such fills are in general avoided is that random strings are assigned very high cost by the soft constraints in our formulation.

The fact that partial solutions can always be extended to satisfy the hard constraints is a difference between the problem being solved by Dr.Fill and those considered elsewhere in the csp literature. Here, however, there is an exception. Much of the work on probabilistic analysis using Markov random fields focuses on a probabilistic maximization similar to ours, once again in an environment where solving the hard constraints is easy but maximizing the score of the result is hard.

A popular inference technique in the Markov setting is *dual decomposition*, where the roles of the variables and values are switched, with Lagrange multipliers introduced corresponding to the variable values and their values then optimized to bound the quality of the solution of the original problem (Sontag, Globerson, & Jaakkola, 2011, for example). This is similar to the csp notion of duality, where the roles of variables and values are also exchanged.

It is not clear how to apply this idea in our setting. In the probabilistic case, the variable values are probabilities selected from a continuous set of real numbers. In the crossword case, the domain is still impractically large but there does not appear to be any natural ordering or notion of continuity between one string value and the next.

There is one further difference between the crossword domain and more standard ones that is also important to understand. Consider the themed crossword called "Heads of State" from the 2010 acpt. The theme entries in this puzzle were common phrases with two-letter state abbreviations appended to the beginning. Thus [Film about boastful jerks?] clues VAIN GLOURIOUS BASTERDS, which is the movie title INGLOURIOUS BASTERDS together with the two-letter state abbreviation VA. [Origami?] clues PAPER FORMING, which is PA adjoined to PERFORMING, and so on.

These multiword fills that do not appear explicitly in the dictionary score fairly badly. In most conventional CSPs, it is reasonable to respond to this by filling the associated words earlier in the search. This allows "better" values to be assigned to these apparently difficult variables. This general idea underlies Joslin and Clements' (1999) "squeaky wheel optimization" and virtually every more recent variable selection heuristic, such as Boussemart et. al's (2004) notion of constraint weighting, and the dom/wdeg heuristic (Lecoutre, Saïs, Tabary, & Vidal, 2009, and others).

In the crossword domain, however, words that score badly in this way should arguably be filled *later* in the search, as opposed to earlier. There is obviously no way for a program such as Dr.Fill, with extremely limited domain knowledge. to figure out that a 21-letter "word" for a [Film about boastful jerks?] should be VAIN GLOURIOUS BASTERDS. The hints suggested by the crossing words are essential (as they are for humans as well). So none of the classic variable selection heuristics can be applied here, and something else entirely is needed. The heuristics that we use are presented in Section 3.

Before moving on, there are two final points that we should make. First, our goal is to find fill for which (4) is maximized; in other words, to maximize our chances of solving the entire puzzle correctly. This is potentially distinct from the goal of entering as many correct words as possible, in keeping with the scoring metric of the ACPT as described in Section 5. The best human solvers generally solve the ACPT puzzles perfectly, however, so if the goal is to *win* the ACPT, maximizing the chances of solving the puzzles perfectly is appropriate.

Second, we designed Dr.Fill so that it could truly exploit the power of its underlying search algorithm. In constructing the $\rho$ function of (5), for example, we don't require that the "correct" solution for a clue $c_i$ be the specific fill $f_i$ for which (5) is minimized, but only hope that the correct $f_i$ be vaguely near the top of the list. The intention is that the hard constraints corresponding to the requirement that the filling words "mesh" do the work for us. As with so many other automated game players (Campbell, Hoane, & Hsu, 2002; Ginsberg, 2001; Schaeffer, Treloar, Lu, & Lake, 1993), we will rely on search to replace understanding.

## 2.4 Data Resources Used by Dr.Fill

One of the most important resources available to Dr.Fill is its access to a variety of databases constructed from online information. We briefly describe each of the data sources here; a summary is in Table 1. The table also includes information on the size of the analogous data source used by Littman's crossword solving program Proverb (Littman et al., 2002).

### 2.4.1 Puzzles

Dr.Fill has access to a library of over 47,000 published crosswords. These include virtually all of the major published sources, including the *New York Times*, the (now defunct) *New York Sun*, the *Los Angeles Times*, *USA Today*, the *Washington Post*, and many others. Most puzzles from the early 1990's on are included.

| data type | source | quantity | quantity (Proverb) |
|-----------|--------|----------|-------------------|
| puzzles | various | 47,693 | 5,142 |
| clues | http://www.otsys.com/clue | 3,819,799 | 350,000 |
| unique clues | http://www.otsys.com/clue | 1,891,699 | 250,000 |
| small dictionary | http://www.crossword-compiler.com | 8,452 | 655,000 |
| big dictionary | various | 6,063,664 | 2,100,000 |
| word roots | http://wordnet.princeton.edu | 154,036 | 154,036 (?) |
| synonyms | WordNet and other online | 1,231,910 | unknown |
| Wikipedia titles | http://www.wikipedia.org | 8,472,583 | — |
| Wikipedia pairs | http://www.wikipedia.org | 76,886,514 | — |

Table 1: Data used by Dr.Fill

Collectively, these puzzles provide a database of just over 3.8 million clues, of which approximately half are unique. This is to be contrasted with the corresponding database in Proverb, which contains some 5,000 puzzles and 250,000 unique clues.

The clue database is available from http://www.otsys.com/clue, a public-domain clue database used by many crossword constructors. The underlying data is compressed, but the source code is available as well and should enable interested parties to decompress the data in question.

### 2.4.2 Dictionaries

As with Proverb, Dr.Fill uses two dictionaries. A small dictionary is intended to contain "common" words, and a larger one is intended to contain "everything". The larger dictionary is an amalgamation from many sources, including Moby[7] and other online dictionaries, Wikipedia titles, all words that have ever been used in crosswords, and so on.

The small dictionary is the "basic English" dictionary that is supplied with Crossword Compiler, an automated tool that can be used to assist in the construction of crosswords.

The large dictionary is much more extensive. Every entry in the large dictionary is also marked with a score that is intended to reflect its crossword "merit". Some words are generally viewed as good fill, while others are bad. As an example, BUZZ LIGHTYEAR is excellent fill. It is "lively" and has positive connotations. The letters are interesting (high Scrabble score, basically), and the combination ZZL is unusual. TERN is acceptable fill; the letters are mundane and the word is overused in crosswords, but the word itself is at least well known. ELIS (Yale graduates) is poor fill. The letters are common, the word is obscure, and it's an awkward plural to boot. Crossword merit for the large dictionary was evaluated by hand scoring approximately 50,000 words (100 volunteers, all crossword constructors, scored 500 words each). The words were then evaluated against many criteria (length, Scrabble score, number of Google hits,[8] appearances in online corpora, etc.) and a linear model was built that best matched the 50,000 hand-scored entries. This model was used to score the remaining words.

---

7. http://icon.shef.ac.uk/Moby/mwords.html

8. I would like to thank Google in general and Mark Lucovsky in particular for allowing me to run the approximately three million Google queries involved here.

Note that the scores here reflect the crossword "value" of the words in isolation, ignoring the clues. Thus we cannot use the dictionaries alone to solve crosswords; indeed, for any particular crossword, there will be many legal fills and the actual solution is unlikely to be anywhere near the "best" fill in terms of word merit alone.

### 2.4.3 GRAMMATICAL AND SYNONYM INFORMATION

Grammatical information is collected from the data provided as part of the WordNet project (Fellbaum, 1998; Miller, 1995). This includes a list of 154,000 words along with their parts of speech and roots (e.g., WALKED has WALK as a root). PROVERB also cites WordNet as a source. In addition, a list of 1.2 million synonyms was constructed from an online thesaurus.

### 2.4.4 WIKIPEDIA

Finally, a limited amount of information was collected from Wikipedia specifically. DR.FILL uses a list of all of the titles of Wikipedia entries as a source of useful names and phrases, and uses a list of every pair of consecutive words in Wikipedia to help with phrase development and fill-in-the-blank type clues. There are approximately 8.5 million Wikipedia titles, and Wikipedia itself contains 77 million distinct word pairs.

## 3. Heuristics

At a high level, most CSPs are solved using some sort of depth-first search. Values are assigned to variables and the procedure is then called recursively. In pseudocode, we might have:

**Procedure 3.1** *To compute* $\texttt{solve}(C, S)$*, a solution to a* CSP $C$ *that extends a partial solution* $S$*:*

```
1  if S assigns a value to every variable in V_C, return S
2  v ← a variable in V_C unassigned by S
3  for each d ∈ D_v(C|_S)
4      do S' ← S ∪ (v = d)
5          C' ← propagate(C|_S')
6          if C' ≠ ∅
7              then Q ← solve(C', S')
8                  if Q ≠ ∅, return Q
9  return ∅
```

We select an unassigned variable, and try each possible value. For each value, we set the variable to the given value and propagate in some unspecified way. We assume that this propagation returns the empty set as a failure marker if a contradiction is discovered, in which case we try the next value for $v$. If the propagation succeeds, we try to solve the residual problem and, if we manage to do so, we return the result.

**Proposition 3.2** *Let $C$ be a* CSP *of size $n$. Then if the* propagate *function is sound, the value* solve$(C, \emptyset)$ *computed by Procedure 3.1 is $\emptyset$ if $C$ has no solutions, and a solution to $C$ otherwise.*

**Proof.** The proof is by induction on $n$. For a CSP of size 1, each live domain value is tried for the variable in question; when one survives the propagate construction, the solution is returned by the recursive call in line 1 and then from line 8 as well.

For larger $n$, if the CSP is not solvable, then every recursive call will fail as well so that we eventually return $\emptyset$ on line 9. If the CSP is solvable, we will eventually set any particular variable $v$ to the right value $d \in D$ so that the recursive call succeeds and a solution is returned. □

For weighted CSPs, the algorithmic situation is more complex because we want to return the best solution, as opposed to any solution. We can augment Procedure 3.1 to also accept an additional argument that, if nonempty, is the currently best known solution $B$. We need the following easy lemma:

**Lemma 3.3** *In a* WCSP *where the costs are non-negative, if $S_1 \subseteq S_2$, then $c(S_1) \leq c(S_2)$.*

**Proof.** This is immediate; more costs will be incurred by the larger set of assignments. Note that this is true for WCSPs generally, not just singly weighted CSPs. □

For convenience, we introduce an "inconsistent" assignment $\perp$ and assume that $c(\perp)$ is infinite. Now we can modify Procedure 3.1 as follows:

**Procedure 3.4** *To compute* solve$(C, S, {}^*B)$, *the best solution to a* WCSP *$C$ that extends a partial solution $S$ given a currently best solution $B$:*

1  **if** $c(S) \geq c(B)$, **return** $B$
2  **if** $S$ assigns a value to every variable in $V_C$, **return** $S$
3  $v \leftarrow$ a variable in $V_C$ unassigned by $S$
4  **for** each $d \in D_v(C|_S)$
5      **do** $S' \leftarrow S \cup (v = d)$
6          $C' \leftarrow$ propagate$(C|_{S'})$
7          **if** $C' \neq \emptyset$, $B \leftarrow$ solve$(C', S', B)$
8  **return** $B$

We use the ${}^*B$ notation at the beginning of the procedure to indicate that $B$ is passed by reference, so that when $B$ is changed on line 7, the value of $B$ used in other recursive calls is changed as well.

In the loop through variable values, we can no longer return a solution as soon as we find one; instead, all we can do is update the best known solution if appropriate. This will, of course, dramatically increase the number of nodes that are expanded by the search. There is some offsetting saving in the comparison on line 1; if the cost of a partial solution is higher than the total cost of the best known solution, Lemma 3.3 ensures that we need not expand this partial solution further. Note that the conditions of the lemma are satisfied in Dr. Fill, since the costs are negated logarithms of probabilities, and the probabilities can be assumed not to exceed one.

**Proposition 3.5** *Let* $C$ *be a* CSP. *Then the value* $\mathtt{solve}(C, \varnothing, \bot)$ *computed by Procedure 3.4 is* $\bot$ *if* $C$ *has no solutions, and the least cost solution to* $C$ *otherwise.*

**Proof.** Suppose first that we drop line 1 and replace line 7 with

$$\textbf{if } C' \neq \varnothing \wedge c(\mathtt{solve}(C', S', B)) < c(B) \textbf{ then } B \leftarrow \mathtt{solve}(C', S', B) \qquad (6)$$

Now the result follows easily by an inductive argument similar to the proof of Proposition 3.1. Every possible solution will be considered, and we will gradually find the least cost one to return.

Consider now Procedure 3.4 as written. If we return a set $S$ on line 2, we must have $c(S) < c(B)$ by virtue of the test on line 1. Thus the new requirement in (6), namely that $c(\mathtt{solve}(C', S', B)) < c(B)$, will always be satisfied and the proof will be complete if we can show simply that the test on line 1 will never discard the best solution. In other words, we need to show that for any solution $S'$ discarded as a result of the test on line 1, we will have $c(S') \geq c(B)$. But this follows directly from Lemma 3.3, since $c(S') \geq c(S)$ and $c(S) \geq c(B)$. □

Procedure 3.4 is the historical method of choice on WCSPs. It is generally referred to as *branch and bound* because the cost of the best solution $B$ found in one branch is used to bound the searches in other branches.

To implement the procedure, we need to specify mechanisms by which variables are selected on line 3 and the domain is ordered on line 4. We discuss value selection first and then variable selection. As described in Section 2.3, the propagation mechanism most useful in crossword solving considers only the direct impact of a word selection on the crossing words.

## 3.1 Value Selection

The performance of Procedure 3.4 depends critically on the order in which values are selected from the domain $D$. The sooner we find good solutions, the earlier we can use the test in line 1 to prune the subsequent search. This *kind* of argument will remain valid even after we replace Procedure 3.4 with other algorithms that are more effective in practice; it is always advantageous to order the search so that the final solution is found earlier, as opposed to later.

There are a variety of elements to this. First, note that all we really need for line 4 of Procedure 3.4 is a function $\mathtt{fill}(v, n)$ that returns the $n$th element of $v$'s live domain $D_v$. On each pass through the loop, we call $\mathtt{fill}$ while gradually increasing the value of $n$. Faltings and Macho-Gonzalez (2005) take a similar approach in their work on "open" constraint programming.

As in the work on open constraint programming, this observation allows us to deal with the fact that not all crossword fills appear explicitly in the dictionary. Our scoring function allows non-dictionary words, but assumes that an apparently unrelated string of words (or of letters) is less likely to be correct than a word or phrase that actually appears in the dictionary. This means that the $\mathtt{fill}$ function can evaluate all of the dictionary possibilities before generating any "multiwords". The multiwords are generated only as needed; by the time they *are* needed, most of the letters in the word are generally filled. This narrows the search for possible multiwords substantially.[9]

The implementation begins by scoring every word of the appropriate length and storing the results with the domain for word $n$. When multiwords are needed and generated, they are added to the end of the domain sets as appropriate.

This approach reduces the value selection problem to two subproblems. First, we need the scoring function $\rho(f_i, c_i)$ that evaluates fill $f_i$ given clue $c_i$. Second, we need to use this scoring function to produce an actual ordering on possible words to enter; the lowest cost word may or may not be the one we wish to try first.

We will not spend a great deal of time describing our scoring function; the details are predictably fairly intricate but the ideas are simple. Fundamentally, we will take the view that has proven so successful elsewhere in computer game players: It is more important that the system be able to search effectively than that it actually have a terribly good idea what it is doing. The power is always more in the search than in the heuristics. This overall search-based approach underlies virtually all of the best computer game players (Campbell et al., 2002; Ginsberg, 2001; Schaeffer et al., 1993) and search-based algorithms have easily outperformed their knowledge-based counterparts (Smith, Nau, & Throop, 1996, for example) in games where direct comparisons can be made.

We implement this idea with a scoring system that is in principle quite simplistic. Words are analyzed based on essentially five criteria:[10]

1. A match for the clue itself. If a clue has been used before, the associated answer is preferred. If a new clue shares a word or subphrase with an existing one, that answer scores well also.

2. Part of speech analysis. If it is possible to parse the clue to determine the likely part of speech of the answer, fill matching the desired part of speech is preferred. The part of speech analysis is based on the WordNet dictionary (Fellbaum, 1998; Miller, 1995), which is then used to search for parse patterns in the clue database. No external syntax or other grammatical mechanisms are used.

3. Crossword "merit" as discussed in Section 2.4.2.

4. Abbreviation. Abbreviations in the dictionary are identified by assuming that words that are generally clued using abbreviations are themselves abbreviations, as described previously. This information is then used in scoring a possible answer to a new clue. What exactly constitutes an "abbreviation" clue is determined by recursively analyzing the clue database.

5. Fill-in-the-blank. Some clues are "fill in the blank"s. These generally refer to a common phrase with a word missing, as in 24-A in Figure 1, where [Line __] clues

---

9. And, as remarked earlier, means that we need to value badly scoring variables *late* in the search as opposed to early.

10. Proverb has some thirty individual scoring modules (Littman et al., 2002), although Littman has suggested (personal communication) that most of the value comes from modules that are analogous to those used by Dr.Fill. Proverb does not analyze the clues to determine the part of speech of the desired fill.

ITEM. These clues are analyzed by looking for phrases that appear in the body of Wikipedia.

These five criteria are then combined linearly. To determine the weights for the various criteria, a specific set of weights $(w_1, \ldots, w_n)$ is selected and then used to solve each of a fixed testbed of puzzles (the first 100 *New York Times* puzzles from 2010). For each puzzle in the testbed, we count the number of words entered by the search procedure before a mistake is made in that the heuristically chosen word is not the one that appears in the known solution to the puzzle. The average number of words entered correctly is then the "score" of $(w_1, \ldots, w_n)$ and the weights are varied to maximize the score.

Given the scoring function $\rho$, how are we to order the values for any particular word? We don't necessarily want to put the best values first, since the value that is best on this word may force us to use extremely suboptimal choices for all of the crossing words.

More precisely, suppose that we assign value $d$ to variable $v$, and that propagation now reduces the variable domains to new values $D_i(C|_{S \cup \{v=d\}})$. An argument similar to that underlying Lemma 2.8 now produces:

**Proposition 3.6** *Let $C$ be an* SWCSP *and $S$ a partial solution, so that $D_u(\pi(C|_{S \cup \{v=f\}}))$ is the domain for $u$ after $v$ is set to $f$ and the result propagated. Then the minimum cost of a solution to $C$ that extends $S \cup \{v = f\}$ is at least*

$$\sum_u \min_{x \in D_u(\pi(C|_{S \cup \{v=f\}}))} \rho(x, u). \qquad \square \qquad (7)$$

We order the variable values in order of increasing total cost as measured by (7), preferring choices that not only work well for the word slot in question, but also minimally increase the cost of the associated crossing words.

This notion is fairly general. In any WCSP, whenever we choose a value for a variable, the choice "damages" the solution to the problem at large; the amount of damage can be determined by propagating the choice made using whatever mechanism is desired (a simplistic approach such as ours, full arc consistency, Cooper's linear relaxation, etc). Cost is incurred not only by the choice just made, but as implied on other variables by the propagation mechanism. (7) says that we want to choose as value for the variable $v$ that value for which the total global cost is minimized, not just the local cost for the variable being valued.

In the crossword domain, this heuristic appears to be reasonably effective in practice. Combined with the variable selection heuristic to be described in the next section, Dr.Fill inserts an average of almost 60 words into a *Times* puzzle before making its first mistake.

## 3.2 Variable Selection

As argued in the previous section, the heuristic we use in valuing a possible fill $f$ for a word slot $s$ in our puzzle is

$$h(f, v) = \sum_u \min_{x \in D_u(\pi(C|_{S \cup \{v=f\}}))} \rho(x, u) - \sum_u \min_{x \in D_u(C|_S)} \rho(x, u) \qquad (8)$$

Because the domain for variable $u$ before setting $v$ to $f$ is $D_u(C|_S)$, the term on the right in (8) gives a lower bound on the best possible score of a complete solution before $v$ is set to $f$ (and this expression is thus independent of $f$).

The value of the term on the left is a lower bound on the best possible score *after* $v$ is set to $f$ because the domain for $u$ after setting $v$ to $f$ and propagating is $D_u(\pi(C|_{S \cup \{v=f\}}))$. The heuristic value of setting $v$ to $f$ is the difference between these two numbers, the total "damage" caused by the commitment to use fill $f$ for variable $v$. Given (8), which variable should we select for valuation at any point in the search?

It might seem that we should choose to value that variable for which $h(f, v)$ is minimized. This would cause us to fill words that could be filled without having a significant impact on the projected final score of the entire puzzle. So we could define the heuristic value of a slot $v$, which we will denote by $H(v)$, to be

$$H(v) = \min_f h(f, v) \tag{9}$$

This apparently attractive idea worked out poorly in practice, and a bit of investigation revealed the reason. Especially early on, when there remains a great deal of flexibility in the choices for all of the variables, there may be multiple candidate fills for a particular clue, all of which appear attractive in that $h(f, v)$ is small. In such a situation, there is really no strong reason to prefer one of these attractive fills to the others, but using (9) as the variable selection heuristic will force us to value such a variable and therefore commit to such a choice.

The solution to this problem is to choose to value not that variable for which $h(f, v)$ is minimized, but the variable for which the *difference* between the minimum value and second-best value is maximal. It is this difference that indicates how confident we truly are that we will fill the slot correctly once we decide to branch on it. If we define min2(S) to be the second-smallest element of a set $S$, then the variable selection heuristic we are proposing is

$$H(s) = \min_f 2\, h(f, v) - \min_f h(f, v) \tag{10}$$

where large values are to be preferred over smaller ones.

As mentioned previously, a combination of (10) and (8) allows Dr. Fill to enter, on average, an initial 59.4 words into a *Times* puzzle before it makes its first error.

It is important to realize that this metric – 59.4 words inserted correctly on average – is *not* because the scoring function accurately places the correct word "first" a large fraction of the time. Instead, our methods are benefiting even at this point from anticipation of how the search is likely to develop; the heuristics themselves are based as much on a glimpse of the future search as they are on the word values in isolation. Indeed, if we use the variable selection described here but switch our value selection heuristic to simply prefer the best fill for the word in question (without considering the impact on subsequent search), the average number of words filled correctly at the outset of the search drops to 25.3, well under half of its previous value.

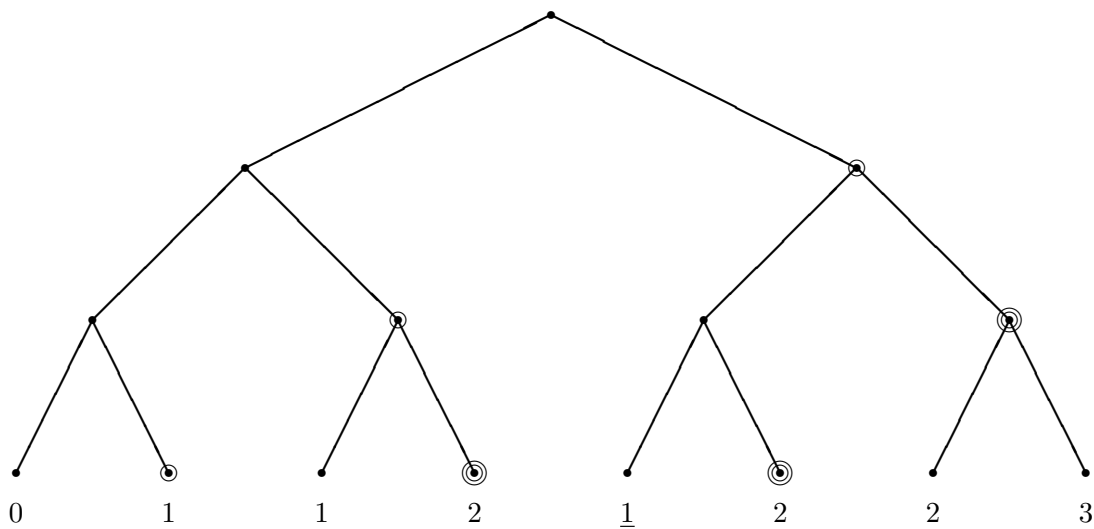

Figure 3: Limited discrepancy search

## 4. Limited Discrepancy Search

Given that DR.FILL can enter nearly sixty correct words in a crossword before making an error, one would expect it to be a strong solver when combined with the branch-and-bound solving procedure 3.4. Unfortunately, this is not the case.

The reason is that this solving procedure suffers from what Harvey (1995) has called the "early mistakes" problem. Once a mistake is made, it impacts the subsequent search substantially and the mistake is never retracted until the entire associated subspace is examined. An initial mistake at depth (say) sixty seems impressive but the quality of the solution below this point is likely to be quite poor, and there is unlikely to be sufficient time to retract the original error that led to the problem.

One way around this problem in CSPs with binary domains is to use *limited discrepancy search*, or LDS (Harvey & Ginsberg, 1995). The idea is that if a heuristic is present, we define the "discrepancy" count of a partial solution $S$ to be the number of times that $S$ violates the heuristic. In Figure 3, we have shown a simple binary search tree of depth three; assuming that the heuristic choice is always to the left, we have labeled each fringe node with the number of times that the heuristic is violated in reaching it.

LDS is an iterative search method that expands the tree using depth-first search and in order of increasing discrepancy count. On the first iteration, only nodes without discrepancies are examined, so the search is pruned at each node in the figure with a single bullseye. On the second iteration, a single discrepancy is permitted and the nodes with double bullseyes are pruned. It is not hard to see that iteration $n$ expands $O(d^n)$ nodes, as the discrepancy limit forms a "barrier" against a full search of the tree. Each iteration also uses only $O(d)$ memory, since the expansion on any individual iteration is depth first. There is some work repeated from iteration to iteration, but since the bulk of the work in iteration $n$ involves nodes that were not expanded in iteration $n - 1$, this rework has little impact on performance. Korf (1996) presents an algorithmic improvement that addresses this issue to some extent.

The point of LDS is that it allows early mistakes to be avoided without searching large portions of the space. In the figure, for example, if the heuristic is wrong at the root of the tree, the node labeled 1 will be explored at the second iteration (with discrepancy limit 1), without the need to expand the left half of the search space in its entirety.

While it is clear that the basic intuition underling LDS is a good match for the search difficulties encountered by Dr.Fill, it is not clear how the idea itself can be applied. One natural approach would be to order the values for any particular word slot, and to then say that using the second value (as opposed to the first) incurred one discrepancy, using the third value incurred two discrepancies, and so on.

This doesn't work. Assuming that the first word in the list is wrong, subsequent words may all score quite similarly. Just because we believed strongly (and wrongly, apparently) that the first word was the best fill does not mean that we have a strong opinion about what to use as a backup choice. The net result of this is that the best solution often uses words quite late in the ordered list; these correspond to a very high discrepancy count and are therefore unlikely to be discovered using this sort of an algorithmic approach.

An alternative idea is to say that a discrepancy is incurred when a variable is selected for branching other than the variable suggested by the variable-selection heuristic (10). This avoids the problem described in the previous paragraph, since we now will pick a fill for a completely different word slot. Unfortunately, it suffers from two other difficulties.

The first (and the less important) is that in some cases, we won't want to change the variable order after all. Perhaps there was a clear first choice and, once that choice is eliminated, there is a clear second choice among the remaining candidate values. In such an instance, we would *want* the "single discrepancy" search choice to try the second fill instead of the first.

More important is the fact that the bad choice is likely to come back on the very next node expansion, when we once again consider the variable in question. The word that looked good when the discrepancy was incurred may well *still* look good, and we will wind up having used the discrepancy but not really having changed the area of the search space that we are considering.

The algorithm that we actually use combines ideas from both of these approaches. As the search proceeds, we maintain a list $P$ of value choices that have been discarded, or "pitched". Each element of $P$ is a pair $(v, x)$ indicating that the value $x$ should not be proposed for variable $v$. The pitched choices remain in the live set, but are not considered as branch values for $v$ until they are forced in that $v$'s live set becomes a singleton. In evaluating the heuristic expressions (8) and (10), pitched values are not considered.

We now incur a discrepancy by pitching the variable and value suggested by the heuristics. Assuming that we then completely recompute both the variable chosen for branching and the value being used, the problems mentioned in the previous paragraphs are neatly sidestepped. We continue to make choices in which we have confidence, and since a pitched value remains pitched as the search proceeds, we do not repeat an apparent mistake later in the search process.

Formally, we have:

**Procedure 4.1** *Let $C$ be a* wcsp. *Let $n$ be a fixed discrepancy limit and suppose that $S$ is a partial solution, $B$ is the best solution known thus far, and $P$ is the set of values pitched in the search. To compute* $\mathtt{solve}(C, S, {}^*B, n, P)$, *the best solution extending $S$ with at most $n$ discrepancies:*

1  **if** $c(S) \geq c(B)$, **return** $B$
2  **if** $S$ assigns a value to every variable in $V_C$, **return** $S$
3  $v \leftarrow$ a variable in $V_C$ unassigned by $S$
4  $d \leftarrow$ an element of $D_v(C|_S)$ such that $(v, d) \notin P$
5  $S' \leftarrow S \cup (v = d)$
6  $C' \leftarrow \mathtt{propagate}(C|_{S'})$
7  **if** $C' \neq \varnothing$, $B \leftarrow \mathtt{solve}(C', S', B, n, P)$
8  **if** $|P| < n$, $B \leftarrow \mathtt{solve}(C, S, B, n, P \cup (v, d))$
9  **return** $B$

**Proposition 4.2** *Let $C$ be a* csp *of size $k$. Then the value* $\mathtt{solve}(C, \varnothing, \bot, n, \varnothing)$ *computed by Procedure 4.1 is $\bot$ if $C$ has no solutions. If $C$ has a solution, there is some $n_0 \leq k(|D|-1)$ such that for any $n \geq n_0$,* $\mathtt{solve}(C, \varnothing, \bot, n, \varnothing)$ *is the least cost solution to $C$.*

**Proof.** There are essentially three separate claims in the proposition, which we address individually.

**1.** If $C$ has no solutions, then the test in line 2 will never succeed, so $B$ will be $\bot$ throughout and the procedure will therefore return $\bot$.

**2.** It is clear that the space explored with a larger $n$ is a superset of the space explored with a smaller $n$ because the test in line 8 will succeed more often. Thus if there is any $n$ for which the best solution is returned, the best solution will also be returned for any larger $n$.

**3.** We claim that for $n = k(|D|-1)$, every solution is considered, and prove this by induction on $k$.

For $k = 1$, we have $n = |D| - 1$. If we are interested in a particular choice $x$ for the unique variable in the problem, then after $|D| - 1$ iterations through line 8, we will either have selected $x$ on line 4 or we will have pitched every other value in which case $x$ will be selected on the last iteration.

The argument in the inductive case is similar. For the variable $v$ selected on line 3, we will use up at most $|D| - 1$ discrepancies before setting the $v$ to the desired value, leaving at least $(n-1)(|D|-1)$ discrepancies to handle the search in the subproblem after $v$ is set. □

**Proposition 4.3** *Let $C$ be a* csp *of size $k$. Then for any fixed $n$, the number of node expansions in computing* $\mathtt{solve}(C, \varnothing, \bot, n, \varnothing)$ *is at most $(k+1)^{n+1}$.*

**Proof.** Consider Figure 4, which shows the top of the lds search tree and labels the nodes with the number of unvalued variables and number of unused discrepancies at each point.

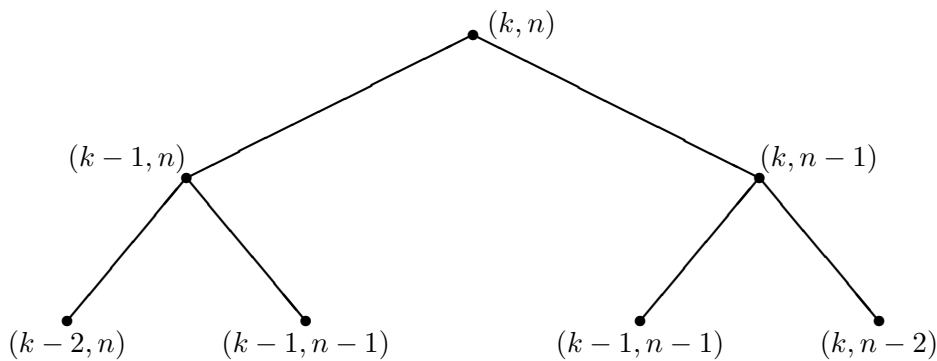

Figure 4: Limited discrepancy search

At the root, therefore, there are $k$ variables left to value and $n$ discrepancies available. If we branch left, we assign a value to some variable. If we branch right, we pitch that choice so that there are still $k$ variables left to value but only $n-1$ discrepancies available.

It follows that if we denote by $f(d, m)$ the size of the search tree under the point with $d$ variables and $m$ discrepancies, we have

$$
\begin{aligned}
f(d, m) &= 1 + f(d, m-1) + f(d-1, m) & (11)\\
&= 1 + f(d, m-1) + 1 + f(d-1, m-1) + f(d-2, m) & (12)\\
&= 2 + f(d, m-1) + f(d-1, m-1) + f(d-2, m) \\
&\vdots \\
&= k + \sum_{i=d-k+1}^{d} f(i, m-1) + f(d-k, m) & (13)\\
&= d + \sum_{i=1}^{d} f(i, m-1) + f(0, m) & (14)\\
&= d + 1 + \sum_{i=1}^{d} f(i, m-1) & (15)\\
&\leq d + 1 + (d-1)[f(d, m-1) - 1] + f(d, m-1) & (16)\\
&= 2 + df(d, m-1)]
\end{aligned}
$$

(11) follows from counting the nodes as in the figure. (12) is the result of expanding the last term in (11), corresponding to expanding the node labeled $(k-1, n)$ in the figure. (13) continues to expand the corresponding term a total of $k$ times, and (14) is just (13) with $k = d$. But $f(0, m) = 1$ because there are no variables left to value, producing (15). (16) follows because $f(i, m-1) \leq f(d, m-1) - 1$ for all $0 < i < d$ (in the figure, every step down the left side is at least one node smaller).

Given $f(d, m) \leq 2 + df(d, m-1)$, we have

$$\frac{f(d, m)}{f(d, m-1)} \leq \frac{2}{f(d, m-1)} + d \leq 1 + d$$

Now $f(d, 0) = d+1$ because the search must progress directly to the fringe if no discrepancies remain. Thus $f(d, m) \leq (1+d)^{1+m}$. Taking $m = n$ and $d = k$ at the root of the tree now produces the desired result. □

## 5. Dr.Fill as a Crossword Solver

At this point, we have described enough of Dr.Fill's underlying architecture that it makes sense to report on the performance of the system as described thus far.[11]

Our overall experimental approach is as follows.

First, we tune the word scoring function $\rho$. Although there are only five basic contributions to the value of $\rho$ for any particular clue and fill, there are currently twenty-four tuning parameters that impact both the five contributions themselves and the way in which they are combined to get an overall value for $\rho$. As described in Section 3, the goal is to maximize the average number of words entered correctly when beginning to solve any of the first 100 *Times* puzzles of 2010.

This tuning process is time consuming; Dr.Fill spends approximately one cpu minute analyzing the clues in any given puzzle to determine the value of $\rho$ for words in its dictionary. This analysis often needs to be repeated if the tuning parameters are changed; it follows that a single run through the testbed of 100 puzzles takes about an hour. The clue analysis is multithreaded and the work is done on an 8-processor machine (two 2.8GHz quad-core Xeons), which reduces wall clock time considerably, but it remains impractical to sample the space of parameter values with other than coarse granularity, and the parameters must in general be tuned independently of one another even though a variety of cross effects undoubtedly exist.

After the tuning is complete, Dr.Fill is evaluated on the puzzles from the 2010 acpt (American Crossword Puzzle Tournament). This is a set of only seven puzzles, but algorithmic and heuristic progress appear to translate quite well into progress on the acpt sample. The puzzles are scored according to the acpt rules, and Dr.Fill's total score is examined to determine where it would have ranked had it been a competitor.

The acpt scoring on any particular puzzle is as follows:

1. 10 points for each correct word in the grid,

2. 25 bonus points for each full minute of time remaining when the puzzle was completed. This bonus is reduced by 25 points for each incorrect letter, but can never be negative.

3. 150 bonus points if the puzzle is solved correctly.

---

11. Dr.Fill is written in C++ and currently runs under MacOS 10.6. It needs approximately 3.5 GB of free memory to run, and is multithreaded. The multithreading uses posix threads and the GUI is written using wxWidgets (www.wxwidgets.org). The code and underlying data can be obtained (for noncommercial use only) by contacting the author. The code can be expected to run virtually unchanged under Linux; Windows will be more of a challenge because Windows has no native support for posix threads.

| version | 1 | 2 | 3 | 4 | 5 | 6 | 7 | total | rank |
|---|---|---|---|---|---|---|---|---|---|
| LDS | 1280 | 925 | 1765 | 1140 | 1690 | 2070 | 1920 | 10790 | 89 (tied) |
| postprocess | 1280 | 1185 | 1790 | 1165 | 1690 | 2070 | 2030 | 11210 | 43 (tied) |
| AND/OR | 1280 | 1185 | 1815 | 1165 | 1690 | 2095 | 2080 | 11310 | 38 |
| best human | 1230 | 1615 | 1930 | 1355 | 1565 | 1995 | 2515 | 12205 | 1 |

Table 2: Results from the 2010 ACPT

Since the puzzles are timed, Dr.Fill needs some sort of termination condition. It stops work and declares its puzzle complete if any of the following conditions occur:

1. A full minute goes by with no improvement in the cost of the puzzle as currently filled,

2. A full LDS iteration goes by with no improvement in the cost of the puzzle as currently filled, or

3. The ACPT time limit for the puzzle is reached.

Results for this and other versions of Dr.Fill appear in Table 2, with scores by puzzle, total score for the tournament, and ranking had Dr.Fill competed. We also give scores for the human (Dan Feyer) who won the event.[12] The first and fourth puzzles are generally the easiest, and the second and fifth puzzles are the hardest. The LDS-based Dr.Fill scored a total of 10,790, good enough for 89[th] place.

After the evaluation is complete, an attempt is generally made to improve Dr.Fill's performance. We examine puzzles from the *Times* testbed (not the ACPT puzzles, which we try to keep as "clean" as possible) and try to understand why mistakes were made. These mistakes can generally be classified as one of three types:

1. Heuristic errors, in that the words entered scored better than the correct ones even though they were not the correct fill,

2. Search errors, where the words entered scored worse than the correct ones but Dr.Fill did not find a better fill because the discrepancy limit was reached, and

3. Efficiency "errors", where points were lost because the search took a long time to complete.

Heuristic errors generally lead to a change in the scoring algorithms in some way, although generally not to the introduction of new scoring modules. Perhaps a different thesaurus is used, or the understanding of theme entries changes. Search errors may lead to modifications of the underlying search algorithm itself, as in Sections 6 and 7. Dr.Fill has a graphical user interface that allows the user to watch the search proceed, and this is often invaluable in understanding why the program performed as it did. Efficiency issues can also (sometimes) be corrected by allowing the visual search to suggest algorithmic modifications; this convinced us that it was worthwhile to treat the overall CSP as an AND/OR tree as discussed in Section 7.

---

12. Feyer went on to win in 2011 as well; Tyler Hinman was the ACPT champion from 2005–2009.

## 6. Postprocessing

An examination of Dr.Fill's completed puzzles based on the algorithms presented thus far reveals many cases where a single letter is wrong, and the problem is with the search instead of the heuristics. In other words, replacing the given letter with the "right" one decreases the total cost of the puzzle's fill. This would presumably have been found with a larger discrepancy limit, but was not discovered in practice.

This suggests that Dr.Fill would benefit from some sort of postprocessing. The simplest approach is to simply remove each word from the fill, and replace it with the best word for the slot in question. If this produces a change, the process is repeated until quiescence.

### 6.1 Formalization and Algorithmic Integration

We can formalize this process easily as follows:

**Procedure 6.1** *Given a* csp *C and a best solution B, to compute* $\mathtt{post}(C, B)$*, the result of attempting to improve B with postprocessing:*

```
 1  change ← true
 2  while change
 3       do change ← false
 4         for each v ∈ C_V
 5             do B' ← B
 6                 unset the value of v in B'
 7                 B' ← solve(C, B', B')
 8                 if c(B') < c(B)
 9                   then B ← B'
10                         change ← true
11  return B
```

We work through the puzzle, erasing each word in line 6. We then re-solve the puzzle (line 7), so that if there is a better choice for that word in isolation, it will be found. If this leads to an improvement, we set a flag on line 10 and repeat the entire process. Note that we only erase one word at a time, since we always begin with the currently best solution in line 5.

As with AC-3, Procedure 6.1 can be improved somewhat by realizing that on any particular iteration, we need only examine variables that share a constraint with a variable changed on the previous iteration. In practice, so little of Dr.Fill's time is spent postprocessing that efficiency here is not a concern.

**Lemma 6.2** *For any* csp *C and solution B*, $c(\mathtt{post}(C, B)) \leq c(B)$. $\qquad\square$

How are we to combine Procedure 6.1 with the basic search procedure 4.1 used by Dr.Fill itself? We can obviously postprocess the result computed by Procedure 4.1 before returning it as our final answer, but if postprocessing works effectively, we should surely postprocess *all* of the candidate solutions considered. That produces:

**Procedure 6.3** *Let $C$ be a* WCSP. *Let $n$ be a fixed discrepancy limit and suppose that $S$ is a partial solution, $B$ is the best solution known thus far, and $P$ is the set of values pitched in the search. To compute* $\texttt{solve}(C, S, *B, n, P)$, *the best solution extending $S$ with at most $n$ discrepancies:*

1   **if** $c(S) \geq c(B)$, **return** $B$
2   **if** $S$ assigns a value to every variable in $V_C$, **return** $\texttt{post}(C, S)$
3   $v \leftarrow$ a variable in $V_C$ unassigned by $S$
4   $d \leftarrow$ an element of $D_v(C|_S)$ such that $(v, d) \notin P$
5   $S' \leftarrow S \cup (v = d)$
6   $C' \leftarrow \texttt{propagate}(C|_{S'})$
7   **if** $C' \neq \emptyset$, $B \leftarrow \texttt{solve}(C', S', B, n, P)$
8   **if** $|P| < n$, $B \leftarrow \texttt{solve}(C, S, B, n, P \cup (v, d))$
9   **return** $B$

The only difference between this and Procedure 4.1 is on line 2, where we postprocess the solution before returning it.

## 6.2 Interaction With Branch and Bound

Further thought reveals a potential problem with this approach. Suppose that our original procedure 4.1 first produces a solution $B_1$ and subsequently produces an improvement $B_2$, with $c(B_2) < c(B_1)$. Suppose also that postprocessing improves both solutions comparably, so that $c(\texttt{post}(B_2)) < c(\texttt{post}(B_1))$. And finally, suppose that postprocessing improves the solutions considerably, so much so, in fact, that $c(\texttt{post}(B_1)) < c(B_2)$.

We are now in danger of missing $B_2$, since it will be pruned by the test on line 1 of Procedure 6.3. $B_2$ will allow us to find a better solution, but only after postprocessing. If we prune $B_2$ early, we will never postprocess it, and the improvement will not be found until a larger discrepancy limit is used.

This suggests that we return to the earlier possibility of postprocessing only the final answer returned by Procedure 4.1, but that may not work, either. Perhaps $B_1$ is improved by postprocessing and $B_2$ is not; once again, the best solution may be lost.

The problem is that branch-and-bound and postprocessing are fundamentally inconsistent; it is impossible to use both effectively. The very *idea* of branch-and-bound is that a solution can be pruned before it is complete if its cost gets too large. The very *idea* of postprocessing is that the final cost of a solution cannot really be evaluated until the solution is complete and the postprocess has been run.

Our "solution" to this is to remove branch and bound from Dr.Fill's search algorithm, producing:

**Procedure 6.4** *Let $C$ be a* WCSP. *Let $n$ be a fixed discrepancy limit and suppose that $S$ is a partial solution, $B$ is the best solution known thus far, and $P$ is the set of values pitched in the search. To compute* $\texttt{solve}(C, S, {}^*B, n, P)$, *the best solution extending $S$ with at most $n$ discrepancies:*

1  **if** $S$ assigns a value to every variable in $V_C$,
2      **return** whichever of $B$ and $\texttt{post}(C, S)$ has lower cost
3  $v \leftarrow$ a variable in $V_C$ unassigned by $S$
4  $d \leftarrow$ an element of $D_v(C|_S)$ such that $(v, d) \notin P$
5  $S' \leftarrow S \cup (v = d)$
6  $C' \leftarrow \texttt{propagate}(C|_{S'})$
7  **if** $C' \neq \emptyset$, $B \leftarrow \texttt{solve}(C', S', B, n, P)$
8  **if** $|P| < n$, $B \leftarrow \texttt{solve}(C, S, B, n, P \cup (v, d))$
9  **return** $B$

Since the test on line 2 ensures that we only change the best solution *$B$ when an improvement is found, all of our previous results continue to hold. But is it really practical to abandon branch and bound as a mechanism for controlling the size of the search?

It is. One reason is that the size of the search is now being controlled by LDS via Proposition 4.3. For any fixed discrepancy limit $n$, this guarantees that the number of nodes expanded is polynomial in the size of the problem being solved.

More important, however, is that experimentation showed that branch-and-bound was ineffective in controlling Dr.Fill's search. The reason is the effectiveness of the (search-anticipating) heuristics used in Dr.Fill itself. These heuristics are designed to ensure that the words inserted early in the search both incur little cost themselves and allow crossing words to incur low cost as well. What happens in practice is that the costs incurred early are extremely modest. Even when a mistake is made, attention typically changes to a different part of the puzzle because filling an additional word $w$ near the mistake begins to have consequences on the expected cost of the words crossing $w$. Eventually, the rest of the puzzle is complete and the algorithm finally begrudgingly returns to $w$ and the cost increases.

Thinking about this, what happens is that while the cost does eventually increase when an error is made, the increase is deferred until the very bottom of the search tree, or nearly so. With so much of the cost almost invariably accumulating at the bottom the search tree, branch and bound is simply an ineffective pruning tool in this domain. The nature of the argument suggests that in other WCSPs that are derived from real-world problems, good heuristics may exist and branch and bound may provide little value in practical problem solving.[13]

### 6.3 Results

The results of Procedure 6.4 appear in Table 2. Dr.Fill's score improves to 11,210, which would have earned it a tie for 43rd place in the 2010 tournament.

---

13. That said, there are certainly real-world problems where branch-and-bound is useful, such as the use of MendelSoft to solve cattle pedigree problems (Sanchez, de Givry, & Schiex, 2008).

## 7. AND/OR Search

There is one further algorithmic improvement that is part of Dr.Fill as the system is currently implemented.

As we watched Dr.Fill complete puzzles, there were many cases where it would fill enough of the puzzle that the residual problem would split into two disjoint subproblems. The search would then frequently oscillate between these two subproblems, which could clearly introduce inefficiencies.

This general observation has been made by many others, and probably originates with Freuder and Quinn (1985), who called the variables in independent subproblems *stable sets*. McAllester (1993) calls a solution technique a *polynomial space aggressive backtracking procedure* if it solves disjoint subproblems in time that is the sum of the times needed for the subproblems independently. Most recently, Marinescu and Dechter (2009) explore this notion in the context of constraint propagation specifically, exploiting the structure of the associated search spaces as AND/OR graphs.

None of this work is directly applicable to Dr.Fill because it needs to be integrated appropriately with LDS. But the integration itself is straightforward:

**Definition 7.1** *Let $C$ be a CSP or WCSP. We will say that $C$ splits if there are nonempty $V_1, V_2 \subseteq V_C$ such that $V_1 \cap V_2 = \varnothing$, $V_1 \cup V_2 = V_C$, and no constraint or weighted constraint in $C$ mentions variables in both $V_1$ and $V_2$. We will denote this as $C = C|_{V_1} + C|_{V_2}$.*

**Proposition 7.2** *Suppose that $C$ is a CSP that splits into $V_1$ and $V_2$. Then if $S_1$ is a solution to $C|_{V_1}$ and $S_2$ is a solution to $C|_{V_2}$, $S_1 \cup S_2$ is a solution to $C$, and all solutions to $C$ can be constructed in this fashion.*

*In addition, if $C$ is a WCSP, then the least cost solution to $C$ is the union of the least cost solutions to $C|_{V_1}$ and $C|_{V_2}$.* □

Note also that we can check to see if $C$ splits in low order polynomial time by checking to see if the constraint graph associated with $C$ is connected. If so, $C$ does not split. If the constraint graph is disconnected, $C$ splits.

**Procedure 7.3** *Let $C$ be a WCSP. Let $n$ be a fixed discrepancy limit and suppose that $S$ is a partial solution, $B$ is the best solution known thus far, and $P$ is the set of values pitched in the search. To compute $\mathtt{solve}(C, S, ^{*}B, n, P)$, the best solution extending $S$ with at most $n$ discrepancies:*

1  **if** $S$ assigns a value to every variable in $V_C$,
2      **return** whichever of $B$ and $\mathtt{post}(C, S)$ has lower cost
3  **if** $C$ splits into $V$ and $W$,
4      **return** $\mathtt{solve}(C|_V, S|_V, B|_V, n, P) \cup \mathtt{solve}(C|_W, S|_W, B|_W, n, P)$
5  $v \leftarrow$ a variable in $V_C$ unassigned by $S$
6  $d \leftarrow$ an element of $D_v(C|_S)$ such that $(v, d) \notin P$
7  $S' \leftarrow S \cup (v = d)$
8  $C' \leftarrow \mathtt{propagate}(C|_{S'})$
9  **if** $C' \neq \varnothing$, $B \leftarrow \mathtt{solve}(C', S', B, n, P)$
10  **if** $|P| < n$, $B \leftarrow \mathtt{solve}(C, S, B, n, P \cup (v, d))$
11  **return** $B$

| Puzzle | Words | Letters | Dr.Fill | | | | Feyer | |
|---|---|---|---|---|---|---|---|---|
| | | | Words wrong | Letters wrong | Time | Score | Time | Score |
| **1** | 78 | 185 | 0 | 0 | 1 | 1280 | 3 | 1230 |
| **2** | 94 | 237 | 8 | 11 | 2 | 1185 | 4 | 1615 |
| **3** | 118 | 301 | 4 | 2 | 2 | 1815 | 6 | 1930 |
| **4** | 78 | 187 | 4 | 2 | 2 | 1165 | 3 | 1355 |
| **5** | 94 | 245 | 0 | 0 | 1 | 1690 | 6 | 1565 |
| **6** | 122 | 289 | 0 | 0 | 1 | 2095 | 5 | 1995 |
| **7** | 144 | 373 | 11 | 13 | 2 | 2080 | 8 | 2515 |
| **total** | 643 | 1817 | 27 | 28 | 11 | 11310 | 35 | 12205 |

Table 3: Results from the 2010 acpt

Note that in line 4, we solve *each* of the split subproblems with a discrepancy limit of $n$. So if (for example) we currently have $n = 3$ with one discrepancy having been used at the point that the split occurs, we will be allowed two additional discrepancies in solving each subproblem, perhaps allowing five discrepancies in total.

In spite of this, the node count will be reduced. If there are $d$ variables remaining when the split is encountered, solving the unsplit problem with $m$ remaining discrepancies might expand $(1 + d)^{1+m}$ nodes (Proposition 4.1), while solving the split problems will expand at most

$$(1 + d_1)^{1+m} + (1 + d - d_1)^{1+m} \tag{17}$$

nodes. A small amount of calculus and algebra[14] shows that $(1+d_1)^{1+m}+(1+d-d_1)^{1+m} \leq (1 + d)^{1+m}$ for $m \geq 1$, so that the split search will be faster even though more total discrepancies are permitted.

The change embodied in Procedure 7.3 significantly improves performance on later lds iterations, and it is arguable that we should exploit this improvement by modifying Dr.Fill's current strategy of terminating the search when an increase in the lds limit does not produce an improved solution. Even without such modification, the increased speed of solution improves Dr.Fill's acpt score by 100 points (one minute faster on puzzles 3 and 6, and two minutes faster on puzzle 7), moving it up to a notional 38[th] place in the 2010 event.

Detailed performance of this final version on the 2010 puzzles is shown in Table 3. For each puzzle, we give the number of words and letters to be filled, and the number of errors made by Dr.Fill in each area. We also give the time required by Dr.Fill to solve the program (in minutes taken), along with the time taken by Dan Feyer, the human winner of the contest. (Feyer made no errors on any of the seven puzzles.) As can be seen, Dr.Fill had 27 incorrect words (out of 643, 95.8% correct) and 28 incorrect letters (out of 1817, 98.5% correct) over the course of the event.

---

14. Differentiating (17) shows that the worst case for the split is $d_1 = 1$, so we have to compare $2^{1+m} + d^{1+m}$ and $(d+1)^{1+m}$. Multiplying out $(d+1)^{1+m}$ produces $d^{1+m} + (1+m)d^m + \cdots$, and $2^{1+m} < (1+m)d^m$ if $d \geq 2$ and $m \geq 1$.

## 8. Related and Future Work

There is a wide variety of work on WCSPs in the academic literature, and we will not repeat any particular element of that work here. What distinguishes our contribution is the fact that we have been driven by results on a naturally occurring problem: that of solving crossword puzzles. This has led us to the following specific innovations relative to earlier work:

- The development of a value selection heuristic based on the projected cost of assigning a value both to the currently selected variable and to all variables with which this variable shares a constraint,

- The development of a variable selection heuristic that compares the difference between the projected cost impacts of the best and second-best values, and branches on the variable for which this difference is maximized,

- A modification of limited discrepancy search that appears to work well for weighted CSPs with large domain sizes,

- The recognition that branch-and-bound may not be an effective search technique in WCSPs for which reasonably accurate heuristics exist, and

- The development and inclusion of an effective postprocessing algorithm for WCSPs, and the recognition that such postprocessing is inconsistent with branch-and-bound pruning.

We do not know the extent to which these observations are general, and the extent to which they are a consequence of the properties of the crossword CSP itself. As discussed previously, crossword CSPs have a relatively small number of variables but almost unlimited domain sizes, and variables whose valuations incur significant cost should in general be filled late as opposed to early.

The two existing projects that most closely relate to Dr.Fill are Proverb (Littman et al., 2002), the crossword solver developed by Littman et. al in 1999, and Watson (Ferrucci et al., 2010), the Jeopardy-playing robot developed by IBM in 2011. All three systems (Watson, Proverb, and Dr.Fill) respond to natural language queries in a game-like setting. In all three cases, the programs seem to have very little idea what they are doing, primarily combining candidate answers from a variety of data sources and attempting to determine which answer is the best match for the query under consideration. This appears to mesh well with the generally accepted view (Manning & Schuetze, 1999) that natural language processing is far better accomplished using statistical methods than by a more classical "parse-and-understand" approach.

The domain differences between Jeopardy and crosswords make the problems challenging in different ways. In one sense, crosswords are more difficult because in Jeopardy, one is always welcome to simply decline to answer any particular question. In crosswords, the entire grid must be filled. On the other hand, the crossing words in a crossword restrict the answer in a way that is obviously unavailable to Jeopardy contestants. Search plays a key role in Dr.Fill's performance in a way that Watson cannot exploit. As a result, Dr.Fill can get by with relatively limited database and computational resources. The program runs on a 2-core notebook with 8 GB of memory and uses a database that is just over 300 MBytes when compressed. WATSON needs much more: 2880 cores and 16 TB of memory. WATSON, like DR.FILL, stores all of its knowledge in memory to improve access speeds – but WATSON relies on much more extensive knowledge than does DR.FILL.

The programs are probably comparably good at their respective cognitive tasks. DR.FILL outperforms all but the very best humans in crossword filling, both in terms of speed (where it is easily the fastest solver in the world) and in terms of accuracy. WATSON, too, outperforms humans easily in terms of speed; its much-ballyhooed victory against human Jeopardy competitors was probably due far more to WATSON's mastery of button pushing than to its question-answering ability. In terms of the underlying cognitive task, WATSON appears to not yet be a match for the best Jeopardy players, who are in general capable of answering virtually all of the questions without error.

DR.FILL itself remains a work in progress. Until this point, we have found heuristic and search errors relatively easily by examining the performance of the program on a handful of crosswords and simply seeing what went wrong. As DR.FILL's performance has improved, this has become more difficult. We have therefore developed automated tools that examine the errors made on a collection of puzzles, identify them as heuristic or search issues, and report the nature of the errors that caused mistakes in the largest sections of fill. The results of these tools will, we hope, guide us in improving DR.FILL's performance still further.

## Acknowledgments

I would like to thank my On Time Systems coworkers for useful technical advice and assistance, and would also like to thank the crossword solving and constructing communities, especially Will Shortz, for their warm support over the years. Daphne Koller, Rich Korf, Michael Littman, Thomas Schiex, Bart Selman, and this paper's anonymous reviewers provided me with invaluable comments on earlier drafts, making the paper itself substantially stronger as a result. The work described in this paper relates to certain pending and issued US patent applications, and the publication of these ideas is not intended to convey a license to use any patented information or processes. On Time Systems will in general grant royalty-free licenses for non-commercial purposes.